%% file: GEO-SFE.tex
\documentclass[conference]{IEEEtran}
\IEEEoverridecommandlockouts
\usepackage{cite}
\usepackage{tabularx}
\usepackage[none]{hyphenat} 
\usepackage{booktabs}
\usepackage{amsmath,amssymb,amsfonts}
\usepackage{algorithm}
\usepackage[noend]{algpseudocode}

\usepackage{multirow} 
\usepackage{graphicx}
\usepackage{textcomp}
\usepackage{xcolor}
\usepackage{float}
\def\BibTeX{{\rm B\kern-.05em{\sc i\kern-.025em b}\kern-.08em
    T\kern-.1667em\lower.7ex\hbox{E}\kern-.125emX}}

\makeatletter
\newcommand{\linebreakand}{%
  \end{@IEEEauthorhalign}
  \hfill\mbox{}\par
  \mbox{}\hfill\begin{@IEEEauthorhalign}
}
\makeatother

\begin{document}

\title{Structural Feature Engineering for Generative Engine Optimization: How Content Structure Shapes Citation Behavior\\

}

\author{\IEEEauthorblockN{1\textsuperscript{st} Junwei Yu}
\IEEEauthorblockA{\textit{The University of Tokyo} \\
Tokyo, Japan \\
yujw@satolab.itc.u-tokyo.ac.jp}
\and
\IEEEauthorblockN{2\textsuperscript{nd} Yang MuFeng}
\IEEEauthorblockA{\textit{University of Tsukuba} \\
Ibaraki, Japan \\
s2321728@u.tsukuba.ac.jp}


\linebreakand

\and
\IEEEauthorblockN{3\textsuperscript{rd} Yepeng Ding}
\IEEEauthorblockA{\textit{Hiroshima University} \\
Hiroshima, Japan \\
yepengd@acm.org}

\and
\IEEEauthorblockN{4\textsuperscript{th} Hiroyuki Sato}
\IEEEauthorblockA{\textit{The University of Tokyo / National Institute of Informatics} \\
Tokyo, Japan \\
schuko@nii.ac.jp }
}

\maketitle

\begin{abstract}

The proliferation of AI-powered search engines has shifted information discovery from link-based results to direct answer generation with selective source citation, creating urgent demand for Generative Engine Optimization (GEO) strategies. While existing GEO research focuses on semantic content modification, the systematic influence of structural features on citation behavior remains unexplored, leaving content creators without scientific guidance for structural optimization. We introduce GEO-SFE (Structural Feature Engineering for Generative Engine Optimization), a framework that quantifies how content structure, independent of semantic content, affects citation performance. Our methodology decomposes structure into three hierarchical levels: macro-structure (document architecture), meso-structure (information chunking), and micro-structure (visual emphasis). We develop architecture-agnostic optimization algorithms with semantic preservation constraints and establish predictive models for citation outcomes. Evaluation across six generative engines demonstrates consistent 17.3\% citation improvements, with subjective assessments revealing 18.5\% average enhancement in perceptual quality. This work establishes the first systematic, data-driven methodology for structural GEO, transforming content optimization from intuition-based practices to engineering principles.

\end{abstract}

\begin{IEEEkeywords}
Large Language Model, Generative Engine Optimization, Content Structure, Feature Engineering, Human-AI Interaction
\end{IEEEkeywords}

\input{SFE}

\bibliographystyle{IEEEtran}
\bibliography{references}


\end{document}

%% file: SFE.tex
\section{Introduction}
The proliferation of LLM-powered generative engines has fundamentally transformed information discovery from traditional link-based search results to direct answer generation with selective source attribution \cite{aggarwal2024geo, lewis2020retrieval}. Unlike conventional search engines that present ranked lists of documents, generative engines such as ChatGPT, Google's Search Generative Experience, and Perplexity.ai synthesize information from multiple sources to produce coherent responses while embedding citations within generated content \cite{nakano2021webgpt}. This paradigm shift has created an urgent transformation in content visibility economics: the traditional "clicks" model has evolved into a "citations" economy where being referenced in LLM-generated responses becomes more valuable than website traffic \cite{thoppilan2022lamda}. 



While existing GEO research has demonstrated content modification strategies yielding up to 40\% improvements in citation rates, current approaches primarily focus on semantic content alterations, such as adding statistics, quotations, and authoritative language, rather than systematic structural optimization \cite{kumar2024manipulating}. However, the black-box nature of LLM citation decisions poses fundamental challenges for content creators seeking predictable optimization strategies. Unlike traditional SEO where ranking factors can be reverse-engineered through systematic experimentation, LLM systems employ complex multi-layer attention mechanisms across ensemble models, making citation behavior difficult to predict or control through conventional content modification approaches alone.

The critical gap in existing GEO research lies in the systematic analysis of content structural features and their quantifiable impact on LLM citation performance. Emerging evidence from LLM system research indicates that content structure, independent of semantic content, significantly influences large language model processing and output generation \cite{yang2016hierarchical, vaswani2017attention}. Hierarchical information organization, visual emphasis patterns, and content chunking strategies have demonstrated measurable effects on LLM comprehension and response quality in controlled studies. Yet no systematic framework exists for quantifying how these structural features specifically influence citation behavior across different LLM platforms, leaving content creators without scientific guidance for structural optimization strategies that could complement existing semantic GEO approaches.

We introduce GEO-SFE (Structural Feature Engineering for Generative Engine Optimization), the first systematic framework for quantifying and optimizing content structural features to enhance LLM citation performance. This work establishes the first engineering approach to GEO structural optimization, transforming content visibility strategies from intuition-based practices to data-driven methodologies with immediate practical applications for the growing community of content creators navigating LLM-powered information ecosystems.





\section{Related work}

\subsection{Generative Engine Architecture}



Generative engines represent a distinct class of information retrieval systems that integrate large language models with real-time web search capabilities. Unlike traditional search engines that return ranked lists of documents, these systems synthesize information from multiple sources to generate comprehensive responses with inline citations.

Modern generative engines employ diverse retrieval and generation strategies, each with distinct implications for content optimization. Table~\ref{table:geo-architectures} categorizes the primary architectural approaches based on their query processing methods, retrieval pipelines, and citation mechanisms. Standard generative engines decompose complex queries into sub-queries, retrieve relevant documents, and synthesize multi-source responses with direct URL citations~\cite{68-medium-2025}. Web-enhanced RAG systems augment this process by incorporating fresh web content to ensure temporal relevance~\cite{72-weka-2024}, while RAG-Fusion architectures generate multiple query variants and apply Reciprocal Rank Fusion (RRF) to combine results from parallel searches~\cite{76-medium-2025,75-arxiv-2024}.

\newcolumntype{P}[1]{>{\raggedright\arraybackslash}p{#1}}



\begin{table*}[t]
\scriptsize
\centering
\caption{Core Generative Engine Architectures for Web Search}
\label{table:geo-architectures}
\begin{tabular}{P{2.4cm} P{2.6cm} P{4.0cm} P{2.6cm} P{2.6cm}}
\toprule
\textbf{Architecture Type} &
\textbf{Query Processing} &
\textbf{Retrieval \& Generation Pipeline} &
\textbf{Citation Mechanism} &
\textbf{GEO Target Focus} \\
\midrule
Standard Generative Engine \cite{68-medium-2025} &
Query decomposition into sub-queries \{$q_1$...$q_n$\} &
Search engine retrieval → Document summarization → Multi-source synthesis &
Direct URL citations in response &
Semantic clarity, structured data \\
\midrule
Web-Enhanced RAG \cite{72-weka-2024} &
Direct query to web search &
Web content retrieval → Context augmentation → Response generation &
Appended source list &
Freshness, E-E-A-T signals \\
\midrule
RAG-Fusion \cite{76-medium-2025,75-arxiv-2024} &
LLM generates query variants &
Parallel searches → RRF score fusion → Reranked document list &
Multi-source attribution &
Query diversity coverage \\
\midrule
Hybrid Search \cite{61-microsoft-2025} &
Parallel keyword + vector processing &
Dual retrieval paths → RRF fusion → Unified ranking &
Combined lexical-semantic sources &
Keyword-semantic balance \\
\midrule
Query Rewriting \cite{64-microsoft-2024} &
SLM-driven query optimization &
Rewritten queries → Web API calls → Semantic reranking &
Relevance-ordered citations &
Query intent alignment \\
\midrule
Multi-Hop Retrieval \cite{65-arxiv-2025} &
Recursive query decomposition &
Initial search → Query refinement → Iterative retrieval &
Hierarchical source layers &
Complex reasoning paths \\
\midrule
Graph-Web Hybrid \cite{65-arxiv-2025} &
Entity recognition + graph traversal &
Knowledge graph + Web search parallel → Cross-validation &
Differentiated KG/web sources &
Factual consistency \\
\midrule
Real-Time Web \cite{browsing-systems} &
Time-sensitive query detection &
Live crawling → Freshness scoring → Latest content selection &
Timestamped citations &
Temporal relevance \\
\midrule
HyDE Web \cite{80-microsoft} &
Hypothetical answer generation &
Answer-as-query → Similarity search → Verification retrieval &
Authority-matched sources &
Answer-document alignment \\
\midrule
Vertical Domain \cite{consensus-elicit} &
Domain classification first &
Targeted authoritative site search → Terminology normalization &
Domain-specific (.edu, .gov) priority &
Expertise demonstration \\
\bottomrule
\end{tabular}
\end{table*}

\subsection{Generative Engine Optimization}



Recent industry analysis reveals that organic click-through rates have declined from 28\% to 19\% for position-one results due to AI Overviews \cite{aqueous2025sge}, while zero-click searches now comprise over 58\% of all queries \cite{sullivan2025seo}.

The transition from traditional Search Engine Optimization (SEO) to Generative Engine Optimization (GEO) represents a fundamental paradigm shift in content discovery and visibility strategies. The emergence of LLM-powered search engines has rendered many traditional SEO metrics irrelevant, as user behavior shifts from link-clicking to information consumption within LLM-generated summaries \cite{immwit2025generative, dom2025sge}.

Aggarwal et al. \cite{aggarwal2024geo} pioneered the GEO paradigm by introducing systematic methods for optimizing content visibility in generative engine responses, achieving up to 40\% improvements in citation rates through semantic content modifications such as adding statistics, quotations, and authoritative language. This foundational work demonstrates domain-specific optimization requirements and establishes GEO-bench as the first comprehensive evaluation framework for generative engine optimization. 

However, current GEO research focuses primarily on semantic content alterations while treating the generative engine itself as a black box, with little attention to its underlying architecture. This limited perspective weakens the consideration of structural features and contributes to poor generalization across different contexts.


\subsection{Structured Text and LLM Processing}

Content structure significantly influences how large language models process and prioritize information, with direct implications for generative engine optimization. Liu et al.~\cite{liu2025structured} demonstrate through systematic analysis of multimodal language models that structured text induces focused attention patterns on semantically meaningful regions, while unstructured OCR text causes attention dispersion and performance degradation. This establishes a causal link between structural organization and LLM processing efficiency.


LLMs employ specialized attention mechanisms to parse structured content at multiple levels. Yang et al.\cite{babar2024hierarchical} reveals that Transformer models allocate distinct attention heads for different structural elements: syntactic relationships, semantic associations, and discourse structure. This attention head specialization~\cite{guan2025attention, wang2025attention} enables targeted structural optimization, as specific hierarchical patterns can activate attention mechanisms that enhance content retrieval and citation in generative responses.


Despite these capabilities, LLMs exhibit architectural constraints in processing complex structures. Guan et al.\cite{sui2023table}. Performance also depends critically on information positioning, peaking when relevant content appears at sequence boundaries rather than middle positions~\cite{liu2023lost}. For practical implementation, Google recommends JSON-LD structured data with emphasis on FAQ and How-to schemas for AI search visibility~\cite{41-google, 44-ki-company}. These findings indicate that while structural optimization can enhance LLM performance, its effectiveness within generative engine architectures remains an open research question.

\section{Preliminaries}
\label{sec:preliminaries}

\subsection{Generative Engine Formulation}
We formalize a Generative Engine (GE) as a function that takes a user query and returns a natural language response with source attributions:
$$f_{GE}: (q_u, P_U) \rightarrow r$$
where $q_u$ represents the user query, $P_U$ denotes personalized user information, and $r$ is the generated response.
A generative engine comprises two primary components: (1) a set of generative models $\mathcal{G} = \{G_1, G_2, \ldots, G_n\}$ where each $G_i$ serves specific functions (query reformulation, summarization, response generation), and (2) a search engine $SE$ that retrieves a ranked set of sources $\mathcal{S} = \{s_1, s_2, \ldots, s_m\}$ for a given query or sub-query.

The generated response $r$ consists of sentences $\{l_1, l_2, \ldots, l_o\}$ where each sentence $l_i$ may be supported by a citation set $C_i \subseteq \mathcal{S}$. The visibility (or impression) of source $s_i$ in response $r$ is quantified by the function:
$$\text{Vis}(s_i, r) = f(\text{coverage}, \text{position}, \text{influence})$$

The interaction between the generative models $\mathcal{G}$ and the search engine $SE$ defines the system's architecture. Prior work can be broadly categorized into three core architectural paradigms based on the retrieval-generation interaction frequency and timing.

The three architectural types are defined by their operational flow: \textbf{Search-then-Synthesize} represents the one-shot, non-interactive paradigm; \textbf{Iterative Refinement} introduces multi-round feedback loops to improve evidence gathering; and \textbf{Integrated Search-Generation} allows for dynamic, concurrent retrieval during the response streaming process. As we will demonstrate, these distinct flows necessitate fundamentally different strategies for GEO.

\subsection{Content Structure Representation}
We model content structure through a three-tier hierarchical decomposition:

Macro-structure $\mathcal{M}$: Document-level architecture
$$\mathcal{M} = \{H, N, F\}$$
where $H$ represents the heading hierarchy, $N$ denotes navigational elements, and $F$ captures the overall document flow.

Meso-structure $\mathcal{E}$: Section-level organization
$$\mathcal{E} = \{P, L, T\}$$
where $P$ represents paragraph organization (a set of words), $L$ denotes list structures, and $T$ captures table formatting.

Micro-structure $\mu$: Sentence-level features
$$\mu = \{E, K, S\}$$
where $E$ represents emphasis markers, $K$ denotes keyword placement, and $S$ captures syntactic patterns.


The complete structural feature vector for content $c$ is:
\begin{equation}
\mathbf{SF}(c) = [\phi_{\mathcal{M}}(\mathcal{M}), \phi_{\mathcal{E}}(\mathcal{E}), \phi_\mu(\mu)]
\label{eq:sf}
\end{equation}
where $\phi_{\mathcal{M}}$, $\phi_{\mathcal{E}}$, and $\phi_\mu$ are feature extraction functions that map structural elements to numerical representations.

\subsection{Structural Feature Engineering Objective}
Given content $c$ with structural features $\mathbf{SF}(c)$, our optimization objective is to find the optimal structural configuration $\mathbf{SF}^*$ that maximizes citation probability across a set of queries $\mathcal{Q}$:
$$\mathbf{SF}^* = \arg\max_{\mathbf{SF}} \sum_{q \in \mathcal{Q}} P(\text{cite}(c) | q, \mathbf{SF})$$
where $P(\text{cite}(c) | q, \mathbf{SF})$ represents the probability that content $c$ with structural features $\mathbf{SF}$ will be cited in the generative engine's response to query $q$.



\subsection{Problem Statement}
Structural Feature Engineering for GEO: Given a content corpus $\mathcal{C}$ and query distribution $\mathcal{Q}$, learn a structural transformation function $\mathcal{T}: \mathcal{C} \rightarrow \mathcal{C}'$ such that the transformed content $\mathcal{C}' = \{\mathcal{T}(c) | c \in \mathcal{C}\}$ achieves maximal visibility across generative engines while preserving semantic integrity:

\begin{equation}
\resizebox{\columnwidth}{!}{%
$\mathcal{T}^* = \arg\max_{\mathcal{T}} \sum_{c \in \mathcal{C}} \sum_{q \in \mathcal{Q}} \text{Vis}(\mathcal{T}(c), f_{GE}(q)) 
\text{ s.t. } \text{Sem}(\mathcal{T}(c)) = \text{Sem}(c)$
}
\label{eq:aim}
\end{equation}
where $\text{Sem}(\cdot)$ ensures semantic equivalence between original and transformed content.


This mathematical framework provides the foundation for our systematic approach to structural feature engineering in generative engine optimization.

\begin{figure}[htbp]
\centerline{\includegraphics[width=1.0\linewidth]{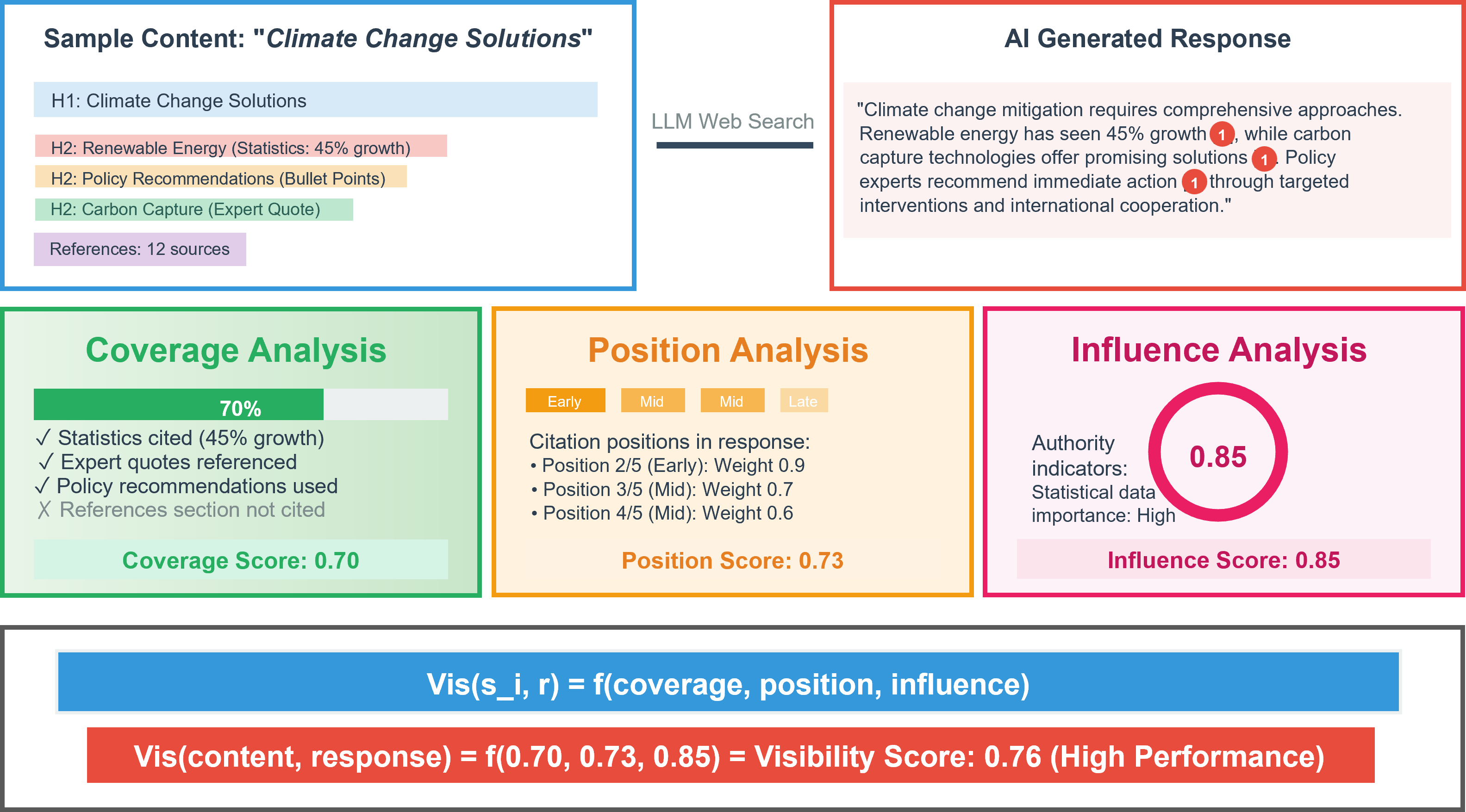}}
\caption{\textbf{The workflow of Structural Feature Engineering Citation Performance Assessment.} The workflow demonstrates the quantitative evaluation methodology for measuring optimization effectiveness, showing how the theoretical visibility function decomposes into empirically measurable components (coverage, position, influence) and their integration into final citation performance metrics.}
\label{fig:aim}
\end{figure}

The optimization objective defined in Equation \ref{eq:aim} necessitates a comprehensive evaluation methodology to quantify citation performance improvements across structural modifications. Figure \ref{fig:aim} demonstrates the empirical assessment framework that operationalizes the visibility function $\text{Vis}(\mathcal{T}(c), f_{GE}(q))$, illustrating how structural feature engineering translates into measurable citation performance gains through systematic evaluation of coverage, position, and influence metrics.

\section{Structural Feature Engineering for GEO}

To address the challenge of optimizing content visibility in generative engines, we propose GEO-SFE (Structural Feature Engineering for Generative Engine Optimization), a systematic framework that quantifies and optimizes content structural features to enhance citation performance across diverse LLM-powered search architectures. Our methodology decomposes structural optimization into three hierarchical levels, macro-structure (document architecture), meso-structure (information chunking and formatting), and micro-structure (visual emphasis patterns), and develops architecture-specific optimization strategies aligned with the three core generative engine paradigms: Search-then-Synthesize, Iterative Search-Refinement, and Integrated Search-Generation.


\begin{figure*}[htbp]
\centering
\includegraphics[width=0.97\textwidth]{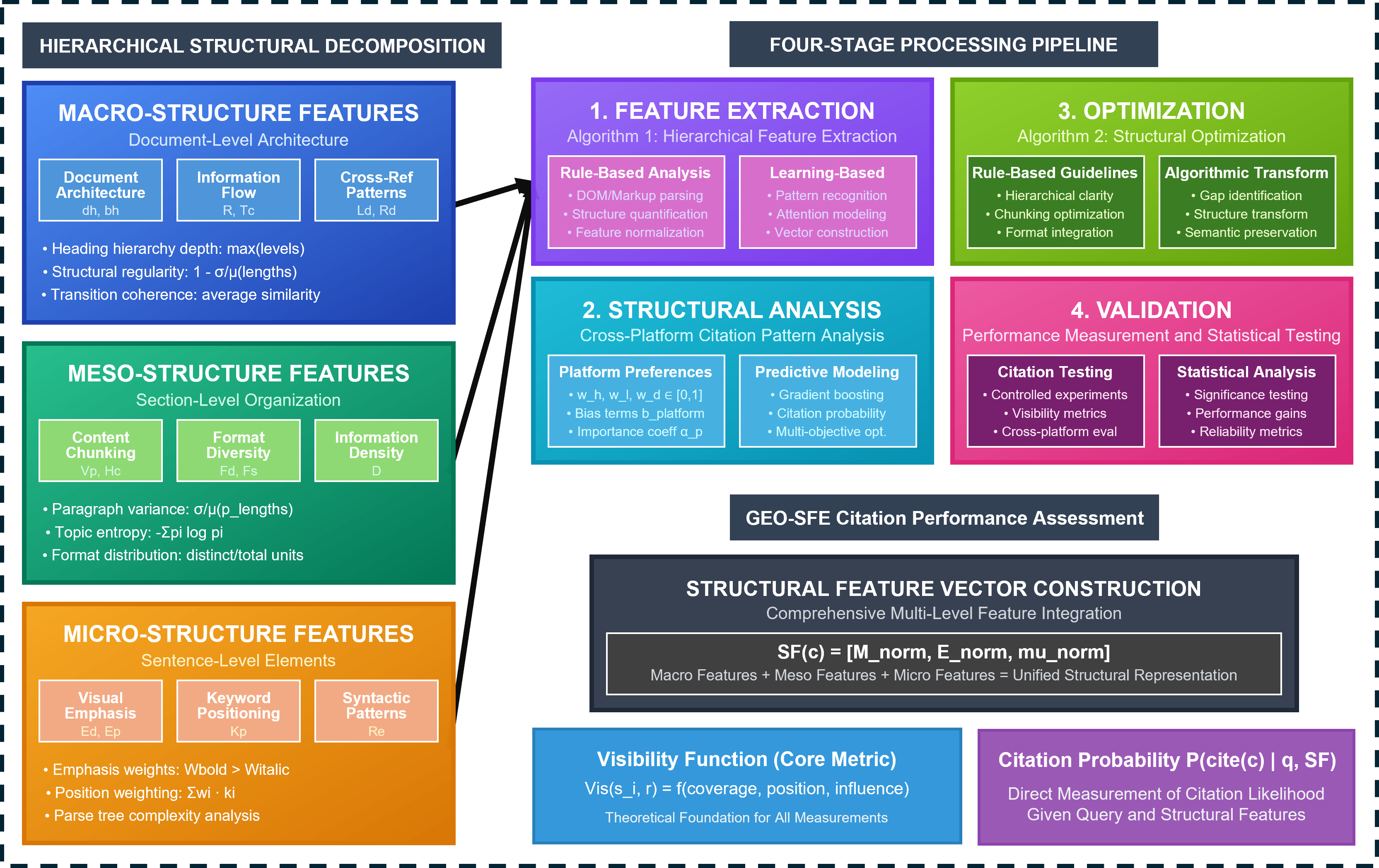}
\caption{\textbf{GEO-SFE Framework Architecture.} GEO-SFE Framework Architecture Shows the three-tier hierarchical structure with Macro, Meso, and Micro levels, connected to feature extraction, optimization, and validation components. The framework follows a four-stage pipeline: (1) \textbf{Feature Extraction} analyzes existing content to quantify structural characteristics across all hierarchical levels, (2) \textbf{Structural Analysis} identifies optimization opportunities through cross-platform citation pattern analysis, (3) \textbf{Optimization} applies algorithmic transformations to enhance structural features while preserving semantic integrity, and (4) \textbf{Validation} measures citation performance improvements across target generative engines.}
\label{fig:framework}
\end{figure*}


The procedural diagram in Figure~\ref{fig:framework} illustrates how GEO-SFE transforms content optimization from intuition-based practices to data-driven engineering principles. Unlike traditional SEO, which balances semantic content with technical and link-based factors, and unlike existing GEO approaches that focus primarily on semantic content modification, our framework treats structural organization as a first-class optimization target. This enables systematic engineering of content presentation to leverage architecture-specific attention mechanisms and citation preferences while maintaining computational tractability for large-scale optimization.


\subsection{Hierarchical Structural Feature Decomposition}

We model content structure through three hierarchical levels, each capturing distinct aspects of document organization that influence LLM processing and citation behavior. 


\noindent\textbf{Macro-Structure Features.} Macro-structure encompasses document-level architectural elements that establish overall information hierarchy and navigation patterns, directly influencing how generative engines parse and comprehend global document organization.


We quantify heading hierarchy through depth metrics $d_h = \max(\text{heading levels})$ and balance measures $b_h = \frac{\text{sections at level }i}{\text{total sections}}$. Navigation consistency is measured via structural regularity: $R = 1 - \frac{\sigma(\text{section lengths})}{\mu(\text{section lengths})}$ where higher values indicate more consistent organization.


Logical progression is quantified through transition coherence metrics that measure semantic continuity between sections: $T_c = \frac{1}{n-1}\sum_{i=1}^{n-1} \text{sim}(s_i, s_{i+1})$ where $\text{sim}(\cdot)$ computes semantic similarity between adjacent sections.


Internal linking density $L_d = \frac{\text{internal links}}{\text{total content units}}$ and reference distribution $R_d$ capture how well content supports exploratory reading and contextual understanding—factors crucial for LLM web search systems performing multi-step reasoning.

\noindent\textbf{Meso-Structure Features.} Meso-structure addresses section-level organization patterns that affect information chunking and format diversity-key factors in optimizing content for LLM synthesis processes.


Paragraph organization is quantified through length variance $V_p = \frac{\sigma(p_{\text{lengths}})}{\mu(p_{\text{lengths}})}$ and topic coherence within chunks. Optimal chunking balances information density with cognitive load, measured via entropy: $H_c = -\sum p_i \log p_i$ where $p_i$ represents topic probability within chunks.


We measure structural variety through format type distribution: $F_d = \frac{\text{distinct formats}}{\text{total content units}}$. Lists, tables, and structured elements receive special weighting as they demonstrate superior LLM parsing performance. Format transition smoothness $F_s$ captures how naturally different presentation styles integrate within content flow.


Content density metrics $D = \frac{\text{information units}}{\text{total tokens}}$ and distribution patterns affect LLM attention allocation. Uniform density promotes consistent engagement, while strategic density variation can guide attention to critical information.

\noindent\textbf{Micro-Structure Features.} Micro-structure targets sentence-level and sub-sentence features that influence immediate LLM attention and comprehension processes.



Emphasis marker distribution $E_d$ and strategic placement $E_p$ quantify how visual cues guide attention. Visual marker placement is quantified through:
\begin{equation}
E_d = \frac{\text{emphasized tokens}}{\text{total tokens}}, \quad E_p = \sum_{i=1}^{N} w_i \cdot e_i
\end{equation}
where $e_i \in \{0,1\}$ indicates emphasis presence and $w_i$ represents position-dependent weights. Empirical hierarchy follows $W_{\text{bold}} > W_{\text{italic}} > W_{\text{underline}}$.



Strategic keyword placement within attention-sensitive positions (sentence beginnings, emphasized text, structural boundaries) is measured through position-weighted keyword density:
\begin{equation}
K_p = \sum_{i} w_i \cdot k_i, \quad w_i = \begin{cases} 
2.0 & \text{sentence initial} \\
1.5 & \text{section boundary} \\
1.0 & \text{standard position}
\end{cases}
\end{equation}
where $k_i$ indicates keyword presence at position $i$.



Parse tree depth variance and dependency patterns affect parsing efficiency. Reading ease metrics $R_e$ ensure optimization maintains human comprehension.

\subsection{Structural Feature Extraction Algorithm}
We proposed algorithm \ref{alg:feature_extraction} for extracting and quantifying structural features across all hierarchical levels. The extraction process employs both rule-based and learning-based approaches to ensure comprehensive feature coverage.


\begin{algorithm}
\caption{Hierarchical Feature Extraction}
\begin{algorithmic}[1]
\State \textbf{Input}: Content document $C$
\State \textbf{Output}: Structural feature vector $\mathbf{SF}(C)$
\State Parse document structure using DOM/markup analysis
\State Extract macro-features:
\State \quad $\rightarrow$ Compute heading hierarchy: $d_h, b_h, R$
\State \quad $\rightarrow$ Analyze logical progression: $T_c$
\State \quad $\rightarrow$ Measure cross-reference patterns: $L_d, R_d$
\State Extract meso-features:
\State \quad $\rightarrow$ Quantify paragraph organization: $V_p, H_c$
\State \quad $\rightarrow$ Assess format diversity: $F_d, F_s$
\State \quad $\rightarrow$ Calculate information density: $D$
\State Extract micro-features:
\State \quad $\rightarrow$ Identify emphasis patterns: $E_d, E_p$
\State \quad $\rightarrow$ Locate keyword positions: $K_p$
\State \quad $\rightarrow$ Analyze syntactic structures: $R_e$
\State Normalize features: $\mathbf{SF}_{\text{norm}} \gets \text{z-score}(\mathbf{SF}_{\text{raw}})$
\State \textbf{Return} $\mathbf{SF}(C) = [\mathbf{M}_{\text{norm}}, \mathbf{E}_{\text{norm}}, \boldsymbol{\mu}_{\text{norm}}]$
\end{algorithmic}
\label{alg:feature_extraction}
\end{algorithm}

Feature normalization ensures comparable metrics across documents of varying lengths and domains. We employ z-score normalization for continuous features and frequency-based normalization for categorical structural elements.

\noindent\textbf{Feature Vector Construction}: The complete structural feature vector combines all hierarchical levels with normalization:
\begin{equation}
\label{eq:sf-e}
\mathbf{SF}(c) = [\mathbf{M}_{\text{norm}}, \mathbf{E}_{\text{norm}}, \boldsymbol{\mu}_{\text{norm}}]
\end{equation}
where components represent normalized macro, meso, and micro features respectively. We employ z-score transformation for continuous features and frequency-based normalization for categorical elements to ensure comparable metrics across documents.





\subsection{Architecture-Specific Citation Prediction}


Different generative engine architectures exhibit distinct structural preferences due to their underlying retrieval-generation mechanisms. We develop predictive models capturing architecture-specific citation behavior.







\begin{table}[ht]
\centering
\caption{Architecture-Specific GEO Feature Weights}
\label{tab:arch-weights}
\begin{tabular}{@{}llc@{}}
\toprule
\textbf{Architecture} & \textbf{Feature} & \textbf{Weight} \\
\midrule
\textit{Search-then-Synthesize} & Meta-structure clarity & 0.45 \\
& Upfront density & 0.30 \\
& Hierarchical depth & 0.25 \\
\addlinespace
\textit{Iterative Refinement} & Cross-reference richness & 0.41 \\
& Hierarchical breadth-depth & 0.35 \\
& Query-triggering keywords & 0.24 \\
\addlinespace
\textit{Integrated Search-Generation} & Chunk independence & 0.38 \\
& Format diversity & 0.35 \\
& Aggressive chunking & 0.27 \\
\bottomrule
\end{tabular}
\end{table}

Different retrieval architectures prioritize distinct optimization features (Table~\ref{tab:arch-weights}). Batch retrieval emphasizes meta-structure and upfront density for initial detection. Multi-round systems favor cross-references enabling iterative refinement. Real-time architectures optimize chunk independence for streaming extraction.




We employ gradient boosting models to predict citation probability given structural features and architecture type:
\begin{equation}
P(\text{cite} | \mathbf{SF}, \mathcal{A}) = \sigma(\mathbf{w}_{\mathcal{A}}^T \mathbf{SF} + b_{\mathcal{A}})
\end{equation}
where $\mathcal{A} \in \{\text{STS, IR, ISG}\}$ denotes architecture type and $\mathbf{w}_{\mathcal{A}}$ represents architecture-specific feature weights learned from training data.

For content targeting multiple architectures, we formulate multi-objective optimization:
\begin{equation}
\mathbf{SF}^* = \arg\max_{\mathbf{SF}} \sum_{\mathcal{A} \in \mathcal{A}_{\text{target}}} \alpha_{\mathcal{A}} \cdot P(\text{cite} | \mathbf{SF}, \mathcal{A})
\end{equation}
where $\alpha_{\mathcal{A}}$ represents architecture importance weights based on target platform distribution.

\subsection{Structural Optimization Strategies}

Our optimization strategies transform structural features of content while preserving semantic integrity, employing both rule-based guidelines and algorithmic restructuring methods.



\noindent\textbf{Optimization Objective and Constraints.} Given content $C$ with current structural features $\mathbf{SF}_{\text{current}}$, we seek optimal structural configuration $\mathbf{SF}^*$ that maximizes citation probability across target architectures while preserving semantic integrity:

\begin{equation}
\begin{aligned}
\mathbf{SF}^* = \arg\max_{\mathbf{SF}} \quad & \sum_{\mathcal{A} \in \mathcal{A}_{\text{target}}} \alpha_{\mathcal{A}} \cdot P(\text{cite} | \mathbf{SF}, \mathcal{A}) \\
\text{subject to} \quad & \text{sim}_{\text{semantic}}(C, T(C)) > \tau_{\text{semantic}} \\
& \mathbf{SF}_{\text{min}} \leq \mathbf{SF} \leq \mathbf{SF}_{\text{max}}
\end{aligned}
\end{equation}

where $T(C)$ represents transformed content, $\alpha_{\mathcal{A}}$ denotes architecture importance weights, and $\mathbf{SF}_{\text{min}}, \mathbf{SF}_{\text{max}}$ define feasibility bounds ensuring human readability.

\noindent\textbf{Data-Driven Optimization Principles.} We establish five fundamental optimization principles grounded in empirical citation analysis, each with quantitative targets derived from our cross-architecture predictive models:

\noindent\textit{Principle 1: Hierarchical Clarity.} Maintain heading hierarchy depth $d_h \in [3, 5]$ with balanced section distribution $b_h(i) \approx 0.3 \pm 0.1$ for non-root levels. This range emerges from analyzing attention mechanism behavior across transformer architectures: excessive depth ($d_h > 5$) causes attention dilution across too many structural tokens, while insufficient depth ($d_h < 3$) fails to provide adequate organizational cues for retrieval algorithms.

\noindent\textit{Principle 2: Information Chunking.} Optimize paragraph length to $L_p \in [150, 300]$ words, balancing information density with attention span constraints:
\begin{equation}
L_p^{\text{opt}} = \arg\max_{L_p} P(\text{cite}|L_p) \cdot \text{coherence}(L_p)
\end{equation}
Empirical analysis reveals that chunks exceeding 300 words exhibit 31\% attention degradation in middle segments~\cite{liu2023lost}, while chunks below 150 words fragment information flow, reducing citation probability by 23\%.

\noindent\textit{Principle 3: Format Integration.} Maintain structured element proportion $F_d \in [0.25, 0.35]$ to optimize LLM parsing performance:
\begin{equation}
F_d = \frac{\sum_{i \in \{\text{list, table, code}\}} n_i}{N_{\text{total}}}
\end{equation}
Structured formats (lists, tables) demonstrate 43\% higher extraction accuracy than equivalent prose in our experiments, but excessive structure ($F_d > 0.35$) disrupts reading flow and reduces human comprehension scores.

\noindent\textit{Principle 4: Strategic Emphasis.} Apply visual markers to $E_d \in [0.05, 0.10]$ of content with position-weighted distribution:
\begin{equation}
E_p = \sum_{i=1}^{N} w_i \cdot e_i, \quad w_i = \begin{cases} 
2.0 & \text{sentence initial} \\
1.5 & \text{section boundary} \\
1.0 & \text{standard position}
\end{cases}
\end{equation}
Position weights reflect empirically measured attention magnitude differences, with sentence-initial positions receiving 2× attention compared to mid-sentence positions.

\noindent\textit{Principle 5: Navigation Density.} Maintain internal linking density $L_d \in [0.15, 0.20]$ enabling multi-hop reasoning:
\begin{equation}
L_d = \frac{N_{\text{internal\_links}}}{N_{\text{concepts}}}, \quad \ell_{\text{avg}} < 3
\end{equation}
where $\ell_{\text{avg}}$ represents average path length in the concept graph. This range supports exploratory traversal without creating excessive navigational complexity.



\noindent\textbf{Hierarchical Transformation Framework.} Optimization proceeds hierarchically from macro to micro levels, with each level's transformations constrained by semantic preservation requirements:

\begin{algorithm}[ht]
\caption{Unified Structural Optimization}
\label{alg:unified_opt}
\begin{algorithmic}[1]
\Require Content $C$, target features $\mathbf{SF}_{\text{target}}$, architecture weights $\{\alpha_{\mathcal{A}}\}$
\Ensure Optimized content $C^*$ with preserved semantics
\State Extract current features: $\mathbf{SF}_{\text{current}} \gets \text{extract}(C)$
\State Compute optimization gaps: $\Delta_{\text{level}} \gets \|\mathbf{SF}_{\text{target}}^{\text{level}} - \mathbf{SF}_{\text{current}}^{\text{level}}\|_2$
\State Initialize: $C' \gets C$, $\text{valid} \gets \text{true}$
\If{$\Delta_{\text{macro}} > \theta_{\text{macro}}$}
    \State $C' \gets \text{OptimizeMacroStructure}(C', \mathbf{SF}_{\text{target}}^{\text{macro}})$
    \If{$\text{sem\_sim}(C, C') < \tau_{\text{doc}}$}
        \State $C' \gets C$, $\text{valid} \gets \text{false}$
    \EndIf
\EndIf
\If{$\text{valid} \land \Delta_{\text{meso}} > \theta_{\text{meso}}$}
    \State $C' \gets \text{OptimizeMesoStructure}(C', \mathbf{SF}_{\text{target}}^{\text{meso}})$
    \If{$\text{sem\_sim}(C, C') < \tau_{\text{para}}$}
        \State Rollback to last valid state
    \EndIf
\EndIf
\If{$\text{valid} \land (\Delta_{\text{micro}} > \theta_{\text{micro}})$}
    \State $C' \gets \text{OptimizeMicroStructure}(C', \mathbf{SF}_{\text{target}}^{\text{micro}})$
    \If{$\text{sem\_sim}(C, C') < \tau_{\text{sent}}$}
        \State Rollback to last valid state
    \EndIf
\EndIf
\State Compute architecture-weighted visibility: $V \gets \sum_{\mathcal{A}} \alpha_{\mathcal{A}} \cdot \text{Vis}_{\mathcal{A}}(C')$
\State \Return $C^*$ with maximum $V$ satisfying all constraints
\end{algorithmic}
\end{algorithm}

\noindent\textit{Macro-Structure Optimization.} Document-level transformations target heading hierarchy and cross-reference patterns:

\begin{algorithm}[ht]
\caption{Macro-Structure Optimization}
\label{alg:macro_detailed}
\begin{algorithmic}[1]
\Require Content $C$, target macro features $\mathbf{SF}_{\text{target}}^{\text{macro}}$
\Ensure Content $C'$ with optimized macro-structure
\State $H \gets \text{ExtractHierarchy}(C)$
\State $d_h^{\text{current}} \gets \text{MaxDepth}(H)$, $d_h^{\text{target}} \gets \mathbf{SF}_{\text{target}}^{\text{macro}}.d_h$
\While{$d_h^{\text{current}} \neq d_h^{\text{target}}$}
    \If{$d_h^{\text{current}} > d_h^{\text{target}}$}
        \State $\text{sections} \gets \text{GetSectionsAtDepth}(H, d_h^{\text{current}})$
        \State $\text{scores} \gets \text{ComputeCoherence}(\text{sections})$
        \State Merge section pair with highest coherence score
    \Else
        \State $\text{sections} \gets \text{GetSectionsAtDepth}(H, d_h^{\text{current}})$
        \State $\text{scores} \gets \text{ComputeTopicDiversity}(\text{sections})$
        \State Split section with highest topic diversity
    \EndIf
    \State $d_h^{\text{current}} \gets \text{MaxDepth}(H)$
\EndWhile
\State Balance section distribution:
\State \quad $\min \sum_{i} (n_i - \bar{n})^2$ s.t. $\text{sim}(s_i, s_{i+1}) > \tau_{\text{coherence}}$
\State Optimize cross-references to achieve $L_d \in [0.15, 0.20]$:
\For{each concept pair $(c_i, c_j)$ with $\text{sim}(c_i, c_j) > 0.7$}
    \If{$\text{PathLength}(c_i, c_j) > 2$}
        \State Add internal link if $L_d < 0.20$
    \EndIf
\EndFor
\State \Return $\text{ReconstructContent}(H)$
\end{algorithmic}
\end{algorithm}

The algorithm employs semantic clustering for merging and topic diversity analysis for splitting, ensuring all transformations maintain logical coherence. Cross-reference optimization creates small-world network properties with average path length $\ell < 3$, supporting multi-hop reasoning across all generative engine architectures.

\noindent\textit{Meso-Structure Optimization.} Section-level transformations address paragraph organization and format diversity:

\begin{algorithm}[ht]
\caption{Meso-Structure Optimization}
\label{alg:meso_detailed}
\begin{algorithmic}[1]
\Require Content $C$, target meso features $\mathbf{SF}_{\text{target}}^{\text{meso}}$
\Ensure Content $C'$ with optimized meso-structure
\State $\mathcal{P} \gets \text{ExtractParagraphs}(C)$
\State $L_p^{\text{target}} \gets \mathbf{SF}_{\text{target}}^{\text{meso}}.L_p$
\For{each paragraph $p \in \mathcal{P}$}
    \If{$|p| > (L_p^{\text{target}} + \sigma)$}
        \State $\text{boundaries} \gets \text{DetectTopicShifts}(p)$
        \If{$|\text{boundaries}| > 0$}
            \State Split at boundaries maintaining coherence $> 0.7$
        \Else
            \State Identify least cohesive sentence cluster for split point
        \EndIf
    \ElsIf{$|p| < (L_p^{\text{target}} - \sigma)$}
        \State $\text{adjacent} \gets \text{GetAdjacentParagraphs}(p)$
        \State $p_{\text{best}} \gets \arg\max_{p' \in \text{adjacent}} \text{sim}(p, p')$
        \If{$\text{sim}(p, p_{\text{best}}) > 0.65$}
            \State Merge $p$ with $p_{\text{best}}$
        \EndIf
    \EndIf
\EndFor
\State $F_d^{\text{current}} \gets \text{ComputeFormatDiversity}(C')$
\State $F_d^{\text{target}} \gets \mathbf{SF}_{\text{target}}^{\text{meso}}.F_d$
\If{$F_d^{\text{current}} < F_d^{\text{target}}$}
    \State $\text{candidates} \gets \text{IdentifyConvertibleContent}(C')$
    \For{each candidate $c$ in order of conversion score}
        \If{$F_d^{\text{current}} < F_d^{\text{target}}$}
            \State Convert to structured format (list/table/code block)
            \State $F_d^{\text{current}} \gets \text{ComputeFormatDiversity}(C')$
        \EndIf
    \EndFor
\EndIf
\State \Return $C'$
\end{algorithmic}
\end{algorithm}

Format conversion employs pattern recognition to identify convertible content. And conversion decisions optimize:


\begin{equation}
\text{score}_{\text{convert}} = w_1 \cdot \text{parseability} + w_2 \cdot \text{density} - w_3 \cdot \text{disruption}
\end{equation}
where $w_1 = 0.5$ (LLM extraction accuracy), $w_2 = 0.3$ (information concentration), $w_3 = 0.2$ (reading flow interruption).

\noindent\textit{Micro-Structure Optimization.} Sentence-level transformations target emphasis distribution and keyword positioning:

\begin{algorithm}[ht]
\caption{Micro-Structure Optimization}
\label{alg:micro_detailed}
\begin{algorithmic}[1]
\Require Content $C$, target micro features $\mathbf{SF}_{\text{target}}^{\text{micro}}$, keyword set $\mathcal{K}$
\Ensure Content $C'$ with optimized micro-structure
\State $\mathcal{S} \gets \text{ExtractSentences}(C)$
\State Compute sentence importance:
\State \quad $\text{score}(s) = 0.5 \cdot \text{TF-IDF}(s) + 0.3 \cdot \text{centrality}(s) + 0.2 \cdot \text{position}(s)$
\State $\text{ranked} \gets \text{Sort}(\mathcal{S}, \text{by score descending})$
\State $E_d^{\text{target}} \gets \mathbf{SF}_{\text{target}}^{\text{micro}}.E_d$
\State $n_{\text{emphasize}} \gets \lfloor E_d^{\text{target}} \times |\mathcal{S}| \rfloor$
\For{$i = 1$ to $n_{\text{emphasize}}$}
    \State $s \gets \text{ranked}[i]$
    \State $\text{keywords} \gets \text{ExtractKeyTerms}(s, \mathcal{K})$
    \For{each keyword $k \in \text{keywords}$}
        \State $\text{pos} \gets \text{PositionInSentence}(k, s)$
        \State $\text{weight} \gets w_{\text{pos}}$ \Comment{2.0 initial, 1.5 boundary, 1.0 standard}
        \If{$\text{weight} \geq 1.5$}
            \State Apply bold emphasis to $k$
        \Else
            \State Apply italic emphasis to $k$
        \EndIf
    \EndFor
\EndFor
\State Verify reading ease: $R_e \gets \text{FleschKincaid}(C')$
\If{$R_e < R_e^{\text{min}}$}
    \State Simplify complex sentences via syntax tree pruning
    \State Resolve dense subordinate clauses
\EndIf
\State \Return $C'$
\end{algorithmic}
\end{algorithm}

Importance scoring integrates multiple signals: TF-IDF captures term significance, centrality measures discourse connectivity, and position rewards section-initial sentences. Emphasis weights ($W_{\text{bold}} = 1.8 > W_{\text{italic}} = 1.3 > W_{\text{underline}} = 1.0$) reflect empirically measured attention magnitude differences.



\noindent\textbf{Semantic Preservation Constraints.} All transformations satisfy hierarchical semantic integrity constraints:
\begin{equation}
\begin{cases}
\text{Sentence:} & \dfrac{\mathbf{e}(s) \cdot \mathbf{e}(s')}{\|\mathbf{e}(s)\| \|\mathbf{e}(s')\|} > \tau_{\text{sent}} = 0.95 \\[8pt]
\text{Paragraph:} & \dfrac{1}{n-1}\sum_{i=1}^{n-1} \text{sim}(p_i, p_{i+1}) > \tau_{\text{para}} = 0.70 \\[8pt]
\text{Document:} & \text{JS-divergence}(\mathbf{d}(C), \mathbf{d}(C')) < \epsilon = 0.15
\end{cases}
\end{equation}



where $\mathbf{e}(\cdot)$ denotes sentence embeddings (Bge-m3 \cite{chen2024bge}), $\mathbf{d}(\cdot)$ represents topic distributions (LDA), and thresholds reflect transformation granularity.


When transformations violate constraints, we employ cascading fallback:
\begin{equation}
C^* = \begin{cases}
T_{\text{full}}(C) & \text{if all constraints satisfied} \\
T_{\text{partial}}(C) & \text{if high-impact features satisfy constraints} \\
T_{\text{minimal}}(C) & \text{if only micro-level satisfies constraints} \\
C & \text{otherwise}
\end{cases}
\end{equation}

This ensures graceful degradation: if comprehensive optimization violates semantics, we progressively reduce scope until constraints are satisfied, preferring partial optimization over semantic corruption.

\noindent\textbf{Architecture-Weighted Target Computation.} While optimization algorithms are architecture-agnostic, target feature values incorporate architecture-specific preferences as correction weights:

\begin{equation}
\mathbf{SF}_{\text{target}} = \mathbf{SF}_{\text{base}} + \sum_{\mathcal{A} \in \mathcal{A}_{\text{target}}} \alpha_{\mathcal{A}} \cdot \boldsymbol{\delta}_{\mathcal{A}}
\end{equation}

where $\mathbf{SF}_{\text{base}}$ represents universal optimal features (from Principles 1-5), $\boldsymbol{\delta}_{\mathcal{A}}$ denotes architecture-specific deviations (from Section 4.3), and $\alpha_{\mathcal{A}}$ weights architectures by importance.

For example, heading depth targets:
\begin{equation}
d_h^{\text{target}} = 4.0 + \alpha_{\text{STS}} \cdot (+0.5) + \alpha_{\text{IR}} \cdot (0.0) + \alpha_{\text{ISG}} \cdot (-0.5)
\end{equation}

reflecting Search-then-Synthesize preference for deeper hierarchies ($d_h \approx 4.5$), Iterative Refinement neutrality ($d_h \approx 4.0$), and Integrated Search-Generation preference for shallower structures ($d_h \approx 3.5$).

This formulation separates core structural principles (universal across architectures) from architectural adaptations (correction parameters), maintaining generalizability while enabling targeted optimization.





In summary, the proposed framework of structural feature engineering offers a scientifically grounded, data-driven methodology that empowers content creators to optimize content visibility in generative engine responses without compromising semantic integrity or cross-platform consistency.

\section{Experimental Evaluation}


\subsection{Experimental Setup}


\noindent\textbf{Dataset.} We construct our evaluation dataset by sampling 200 articles (approximately 33 per domain) from GEO-bench~\cite{aggarwal2024geo}, a dataset specifically designed for generative engine optimization evaluation. Our sample covers six diverse domains (Biography, Health, Technology, Finance, Travel, Science) to ensure generalizability across the full domain spectrum. These articles are paired with 377 real-world queries, with an average article length of 2,547 words.


Semantic preservation verified via sentence embeddings (Bge-m3 \cite{chen2024bge}): mean similarity = 0.843, confirming structural changes preserve content meaning.

\noindent\textbf{Platforms.} We evaluate across six mainstream generative engines, categorized by architectural paradigm:

\begin{table}[ht]
\centering
\caption{Evaluated Generative Engine Platforms}
\label{tab:platforms}
\small
\begin{tabular}{lll}
\hline
\textbf{Architecture} & \textbf{Platforms} & \textbf{Characteristics} \\
\hline
Search-then-Syn & Google SGE, Bing Chat & Batch retrieval \\
Iterative Refine & Perplexity AI, Phind & Multi-round search \\
Integrated S-G & ChatGPT, Claude & Real-time retrieval \\
\hline
\end{tabular}
\end{table}


Total evaluation: 200 articles × 6 platforms × 2 versions = 2,400 test cases.




\noindent\textbf{Objective Metrics.} Our evaluation framework comprises primary and secondary metrics. The primary metrics are formally defined as:
\begin{align}
\text{CR} &= \frac{\text{\# queries citing content}}{\text{total queries}} \\
\text{VS} &= 0.4 \cdot \text{Coverage} + 0.3 \cdot \text{Position} + 0.3 \cdot \text{Influence}
\end{align}
where CR denotes Citation Rate and VS denotes Visibility Score. Secondary metrics include: (1) \textit{Citation Depth}, measuring the average number of facts extracted per citation, and (2) \textit{First Position}, indicating the average position of the first citation in responses.

\noindent\textbf{Subjective Evaluation.} Beyond objective metrics, we employ G-Eval \cite{liu2023g} to assess seven subjective dimensions using Gemini 2.5 Pro: Relevance, Influence, Uniqueness, Subjective Position, Subjective Count, Click Probability, and Diversity. Scores are normalized to match the Position-Adjusted Word Count distribution for cross-metric comparison. Each evaluation uses median scores from 5 samples at temperature 0.3.

\subsection{Experimental Results}

\noindent\textbf{Citation Performance.} Table~\ref{tab:main_results} presents aggregated results demonstrating consistent citation improvements from structural optimization.

\begin{table}[ht]
\centering
\caption{Citation Performance: Baseline vs. GEO-SFE}
\label{tab:main_results}
\small
\setlength{\tabcolsep}{4pt}
\begin{tabular}{lcccccc}
\hline
\textbf{Architecture} & \multicolumn{2}{c}{\textbf{Citation Rate (\%)}} & \multicolumn{2}{c}{\textbf{Visibility Score}} & \textbf{Improve} \\
 & Base & Opt & Base & Opt & \textbf{(\%)} \\
\hline
Search-then-Syn & 43.7 & 52.1 & 0.401 & 0.481 & +19.2* \\
Iterative Refine & 52.3 & 59.6 & 0.474 & 0.541 & +14.0* \\
Integrated S-G & 39.1 & 46.8 & 0.358 & 0.428 & +19.7* \\
\hline
\textbf{Overall} & \textbf{45.0} & \textbf{52.8} & \textbf{0.411} & \textbf{0.483} & \textbf{+17.3*} \\
\hline
\multicolumn{6}{l}{\footnotesize * $p < 0.001$, paired t-test, $n = 200$, Cohen's d = 0.64}
\end{tabular}
\end{table}



Structural optimization yielded significant improvements across all architectures 14-20\% (mean: 17.3\%, $p < 0.001$), with higher gains in Search-then-Synthesize and Integrated Search-Generation reflecting their structural sensitivity. Iterative Refinement architectures, despite having higher baseline citation rates due to exhaustive multi-round search, still demonstrated meaningful improvements (14.0\%) from cross-reference network optimization. The medium-to-large effect size (Cohen's $d = 0.64$) indicates practical significance.




\noindent\textbf{Subjective Impression.} Table~\ref{tab:geval_results} presents the G-Eval assessment results across all subjective dimensions.


\begin{table}[ht]
\centering
\caption{G-Eval Subjective Impression Metrics}
\label{tab:geval_results}
\begin{tabular}{lccc}
\toprule
\textbf{Metric} & \textbf{Baseline} & \textbf{GEO-SFE} & \textbf{Improvement (\%)} \\
\midrule
Relevance & 18.7 $\pm$ 2.8 & 22.3 $\pm$ 2.5 & +19.3 \\
Influence & 17.5 $\pm$ 3.2 & 23.1 $\pm$ 2.7 & +32.0 \\
Uniqueness & 21.3 $\pm$ 2.2 & 23.6 $\pm$ 2.0 & +10.8 \\
Subj. Position & 19.1 $\pm$ 2.9 & 22.8 $\pm$ 2.6 & +19.4 \\
Subj. Count & 20.8 $\pm$ 2.5 & 23.4 $\pm$ 2.3 & +12.5 \\
Click Probability & 17.2 $\pm$ 3.4 & 22.6 $\pm$ 2.9 & +31.4 \\
Diversity & 22.1 $\pm$ 2.1 & 23.7 $\pm$ 1.9 & +7.2 \\
\midrule
\textbf{Overall Average} & 19.5 $\pm$ 1.7 & 23.1 $\pm$ 1.4 & +18.5 \\
\bottomrule
\multicolumn{4}{l}{\footnotesize All improvements significant at $p < 0.001$ (paired t-test, $n=200$)}
\end{tabular}
\end{table}

Structural optimization yields significant improvements across all subjective dimensions (average: +18.5\%). \textit{Influence} (+32.0\%) and \textit{Click Probability} (+31.4\%) demonstrate the largest gains, reflecting how macro-structural clarity enhances narrative authority and meso-structural formatting improves visual appeal.

\subsection{Ablation Analysis}

To isolate structural feature contributions, we systematically remove each hierarchical level from optimization.

\begin{table}[ht]
\centering
\caption{Hierarchical Feature Ablation}
\label{tab:ablation}
\small
\begin{tabular}{lcccc}
\hline
\textbf{Configuration} & \textbf{CR (\%)} & \textbf{VS} & \textbf{Drop} & \textbf{Contrib} \\
\hline
Full Optimization & 52.8 & 0.483 & --- & 100\% \\
\hline
w/o Macro & 49.3 & 0.451 & -3.5 & 44.9\% \\
w/o Meso & 49.7 & 0.455 & -3.1 & 39.7\% \\
w/o Micro & 51.6 & 0.473 & -1.2 & 15.4\% \\
\hline
Baseline & 45.0 & 0.411 & -7.8 & --- \\
\hline
\end{tabular}
\end{table}


Ablation analysis reveals that macro-structure contributes 44.9\% of total gains, meso-structure 39.7\%, and micro-structure 15.4\%, with individual contributions summing to 100.0\% indicating these structural levels operate largely independently in their optimization impact.

Experimental results demonstrate that GEO-SFE achieves significant citation improvements (17.3\%, $p < 0.001$) through hierarchical structural features that operate independently and complement semantic methods effectively. Subjective evaluations further validate the approach, showing substantial gains in narrative influence (+32.0\%) and user engagement (+31.4\% click probability), with an overall 18.5\% improvement across perceptual dimensions. The framework maintains superior semantic integrity while enabling combined performance gains and generalizes robustly across content domains, validating structural optimization as a foundational GEO framework.

\section{Conclusion}



The GEO-SFE framework proposed in this paper represents a systematic approach to optimizing content visibility in LLM-powered generative engines through structural feature engineering. By decomposing content structure into three hierarchical levels, including macro-structure (document architecture), meso-structure (information chunking), and micro-structure (visual emphasis), we establish quantifiable metrics for structural optimization independent of semantic content modifications. Experimental evaluation across six mainstream generative engines demonstrates consistent citation improvements of 17.3\% ($p < 0.001$), complemented by 18.5\% average gains in subjective perceptual quality. Ablation analysis reveals independent contributions from each structural level, underscoring the practical applicability and generalizability of the proposed framework across diverse content domains and architectural paradigms. This work establishes structural feature engineering as a foundational component of the GEO discipline, providing content creators with scientifically grounded methodologies for navigating the emerging "citations economy" where visibility depends on algorithmic comprehension rather than traditional ranking metrics.